\documentclass{article}
\usepackage{spconf,amsmath,graphicx}
\usepackage{amssymb}
\usepackage{array}
\usepackage{hhline, booktabs}
\usepackage{subcaption}
\usepackage{multirow}

\title{Weight-based Mask for Domain Adaptation}
\name{Eunseop Lee \quad Inhan Kim \quad Daijin Kim}
\address{Department of Computer Science and Engineering, 
 Pohang University of Science and Technology, Korea}
\begin{document}
%
\maketitle
\begin{abstract}
In computer vision, unsupervised domain adaptation (UDA) is an approach to transferring knowledge from a label-rich source domain to a fully-unlabeled target domain. Conventional UDA approaches have two problems. The first problem is that a class classifier can be biased to the source domain because it is trained using only source samples. The second is that previous approaches align image-level features regardless of foreground and background, although the classifier requires foreground features. To solve these problems, we introduce Weight-based Mask Network (WEMNet) composed of Domain Ignore Module (DIM) and Semantic Enhancement Module (SEM). DIM obtains domain-agnostic feature representations via the weight of the domain discriminator and predicts categories. In addition, SEM obtains class-related feature representations using the classifier weight and focuses on the foreground features for domain adaptation. Extensive experimental results reveal that the proposed WEMNet outperforms the competitive accuracy on representative UDA datasets.
\end{abstract}

\section{Introduction}
\label{sec:intro}

Deep neural networks have been improved to achieve high performance in the field of computer vision tasks. However, when the deep neural networks are directly applied to new task-specific datasets, their performance degrades because of data bias or domain shift \cite{torralba2011unbiased}. Also, the manual labeling of a large number of high-quality ground-truths is expensive and time-consuming. 

To avoid this cost, unsupervised domain adaptation (UDA) approaches have been introduced. It transfers the knowledge from a label-rich source domain to fully-unlabelled target domain. The traditional deep UDA methods \cite{ganin2015unsupervised, tzeng2017adversarial, zhu2022localized, quinonero2008covariate, long2017deep} attempt to learn domain-invariant models that bridge the source and target domains. Adversarial domain generalization (ADG) approaches \cite{ganin2015unsupervised, tzeng2017adversarial, zhu2022localized} use domain discriminators to minimize the domain adversarial objectives and the gap between the source and target distributions. Other methods aim to reduce the domain discrepancy between the source and target domains by using metrics, such as Maximum Mean Discrepancy (MMD) \cite{quinonero2008covariate} or joint MMD (JMMD) \cite{long2017deep}. All previous UDA methods have achieved remarkable performance, mostly focusing on generating transferable deep representations by reducing cross-domain distribution discrepancies. However, two critical problems are yet be solved. The first is \textit{classification bias problem}. UDA approaches assume that if both domain features can be mapped to a common feature space, these domain-invariant features can be applied to a classifier trained only from the source domain. However, the classifier is biased toward the source domain, which can negatively affect the target samples. The second problem is \textit{background alignment problem}. The preceding methods attempt to align image-level features for the source and target domains. Domain alignment is conducted regardless of foreground or background. Although background features do not contain any class discriminability, these methods can align the corresponding background features.

\begin{figure}[]{}
     \centering
     \includegraphics[width=\linewidth]{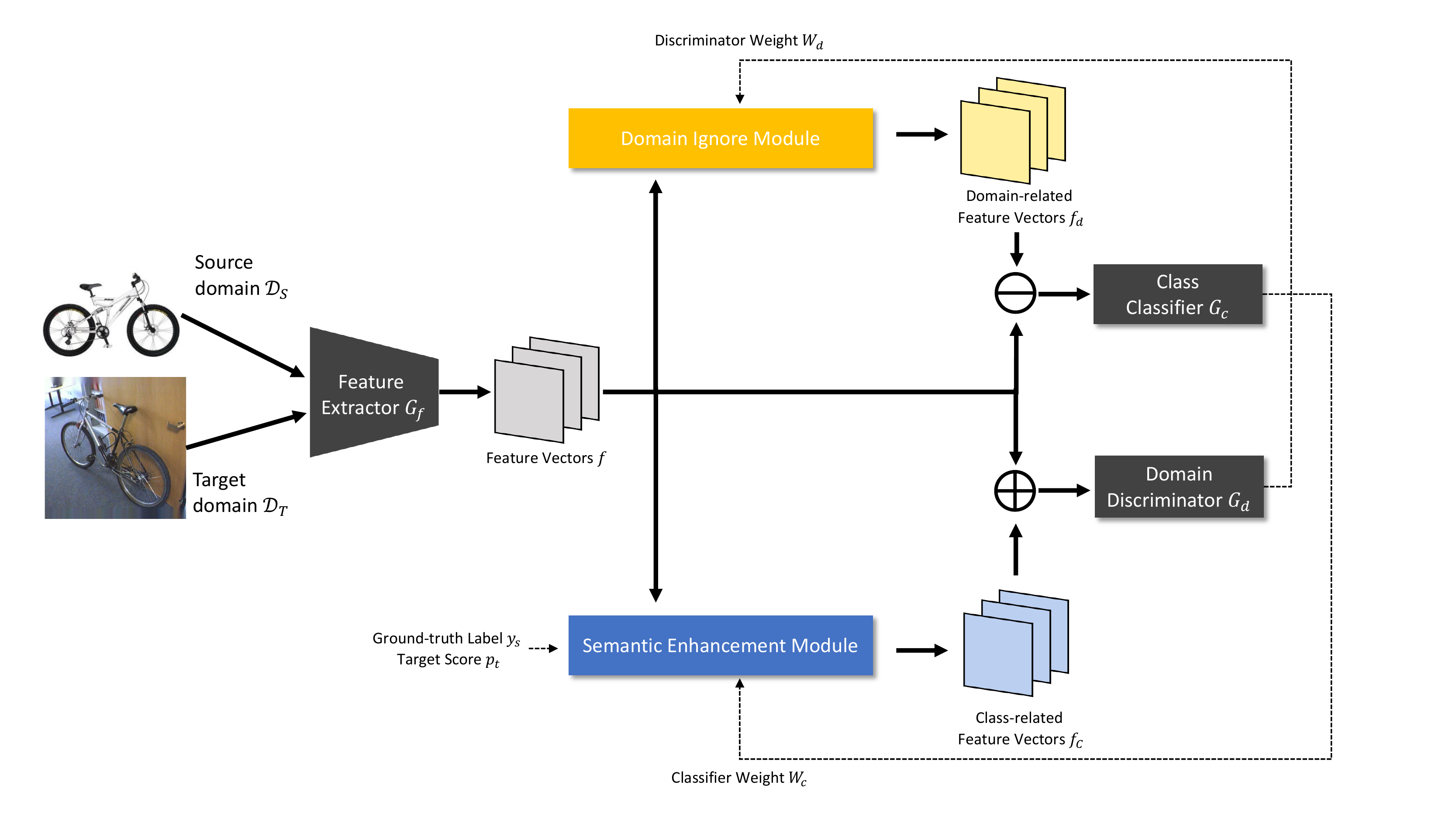}
     \caption{Overall structure of the proposed WEMNet. It consists of two modules: Domain Ignore Module (DIM) and Semantic Enhancement Module (SEM).}
     \label{fig:overall}
\end{figure}

To solve these problems, we propose Weight-based Mask Network (WEMNet) composed of Domain Ignore Module (DIM) and Semantic Enhancement Module (SEM). 
To prevent the class classifier from being biased to the source domain, DIM removes the domain-related information from the extracted features using domain mask and predicts categories based on the corresponding domain-agnostic feature vectors. SEM enhances the foreground features for domain adaptation using class-related features. To obtain these features, we use class mask to keep parts that are class-related based on the classifier weight. Because the proposed method reuses the weights of the existing classifier and domain discriminator, it does not require much additional parameters or computations, and it can be flexibly applied in any ADG framework. To demonstrate the flexibility of the proposed method, we adapt representative frameworks (DANN \cite{ganin2015unsupervised}, and MSTN \cite{xie2018learning}) as baselines. Comprehensive experiments are conducted using popular benchmark datasets (Office-31 \cite{saenko2010adapting}, and VisDA-2017 \cite{peng2017visda}).

\section{Weight-based Mask Network}
\label{sec:method}
As depicted in Fig.\ref{fig:overall}, the proposed WEMNet consists of DIM and SEM. Both modules follow the concept of Weight-based Mask $M$. $M$ is derived from the weight $W=[w_1, w_2, ..., w_{N_h}]$ of the domain discriminator $G_d$ for DIM and of the classifier $G_c$ for SEM. Concretely, let $f \in \mathbb{R}^{N_b \times N_h}$ be the output of the backbone $G_f$, where $N_b$ and $N_h$ are the batch size and channel size, respectively. Both $G_d$ and $G_c$ consist of a fully-connected layer, $W \cdot f$, where $W$ is trained to consider meaningful information. If $w_i \in W$ is the largest or smallest element, it is the most important factor in determining the domain or class label. Therefore, we assume that $W$ can obtain domain- or class-related features from $f$. Based on this concept, DIM removes domain-related features $f_d$ using domain mask $M_d$, and SEM uses class mask $M_c$ to reinforce class-related features $f_c$ for domain adaptation.


\subsection{Task Formulation}
In the task of UDA, we are given a labeled source domain $\mathcal{D}_S = {(x^i_s, y^i_s)}_{i=1}^{N_s}$, where $x^i_s$ is the $i$-th image and $y^i_s$ is its category label. In addition, we take unlabeled target domain $\mathcal{D}_T = {(x^i_t)}_{i=1}^{N_t}$. The source and target domains are drawn from different joint probability distributions $P(X_s, Y_s)$ and $Q(X_t, Y_t)$ respectively (i.e. $P \neq Q$), but the categories are exactly the same. The goal is to transfer the knowledge in $\mathcal{D}_S$ to $\mathcal{D}_T$ and to increase the classification performance.

\begin{figure}
     \centering
     \begin{subfigure}[t]{0.4\textwidth}
         \centering
         \includegraphics[width=\textwidth]{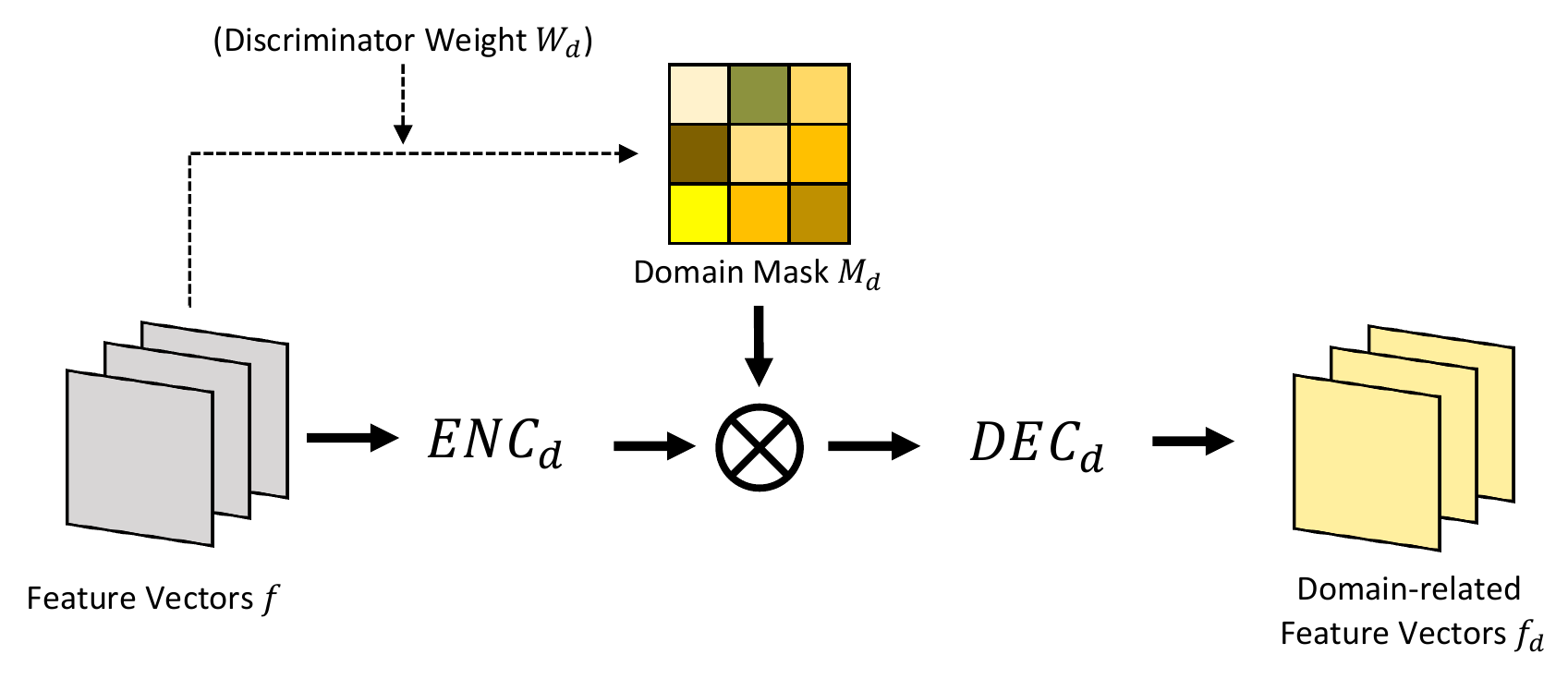}
         \caption{Domain Ignore Module}
         \label{subfig:dim}
     \end{subfigure}
     \begin{subfigure}[b]{0.4\textwidth}
         \centering
         \includegraphics[width=\textwidth]{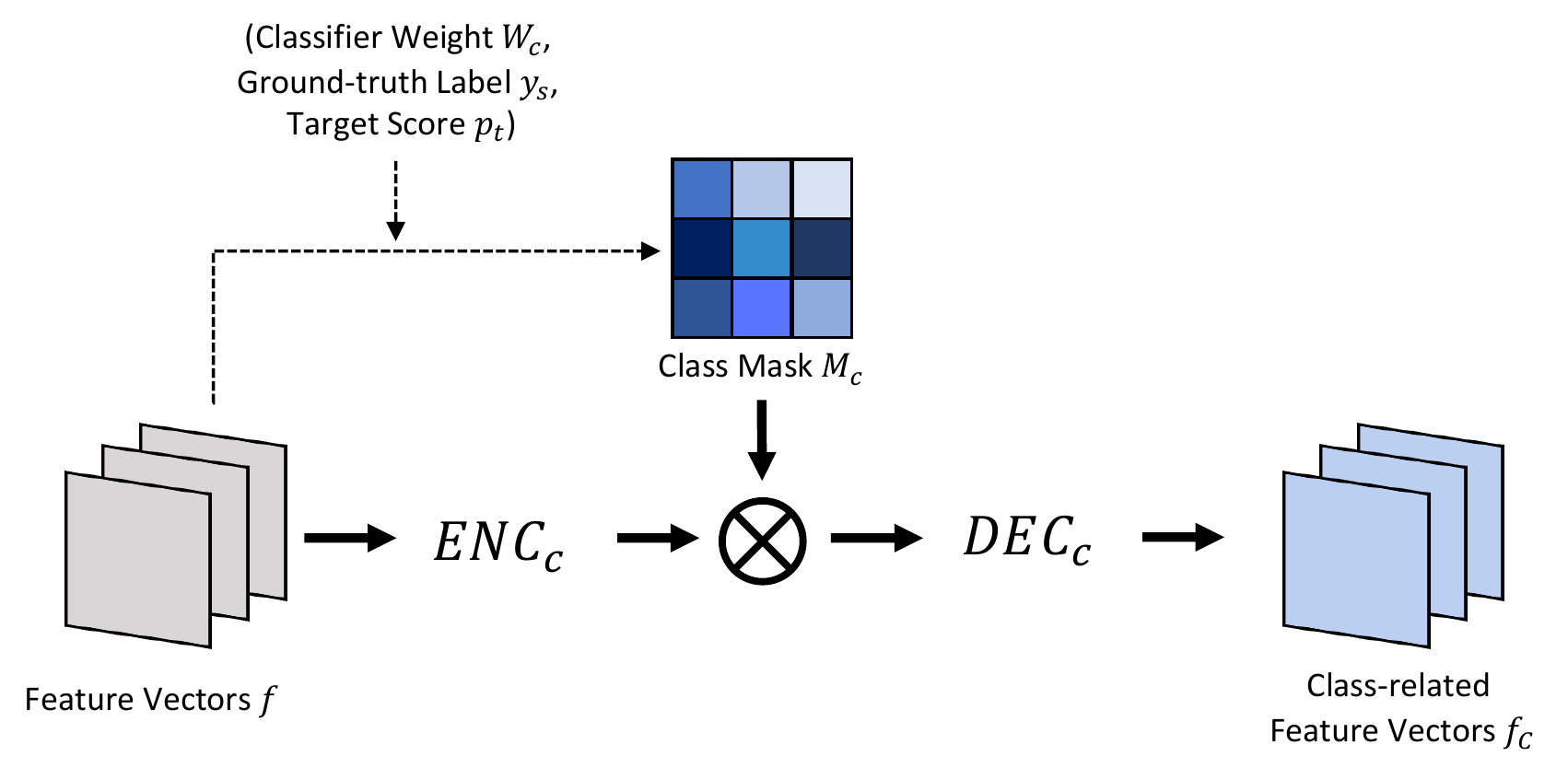}
         \caption{Semantic Enhancement Module}
         \label{subfig:sem}
     \end{subfigure}
        \caption{(a): Domain Ignore Module. (b): Semantic Enhancement Module.}
        \label{fig:dim_sem}
\end{figure}

\subsection{Domain Ignore Module}
\label{subsec:dim}
Because $G_c$ is trained using only the source samples, it can be biased toward the source domain. This \textit{classification bias problem} causes misclassification of the target feature vectors and degrades the classification accuracy. 

To solve this problem, DIM eliminates $f_d$ from $f$, and predicts categories by using the domain-agnostic features $\hat{f}_d$ as shown Fig. \ref{subfig:dim}. Specifically, we construct the domain threshold $T_d$ by averaging the $|W_d|$ across the channel dimension and then filter out the weight elements depending on $T_d$. The remaining elements are the keypoints to distinguish domains. With the unfiltered values, we define $M_d \in \mathbb{R}^{1 \times N_h}$ as,

\begin{equation}
T_d = \frac{1}{N_h} \sum \sigma(|W_d|) 
\end{equation}

\begin{equation}
M_d = I(\sigma|W_d| > T_d),
\end{equation}

\noindent where $\sigma(\cdot)$ is the sigmoid function, and $I(\cdot)$ is the indication function.

Using $M_d$, we construct $\hat{f}_d$ by multiplying $M_d$ by $f$. Subsequently, we remove domain-related information from $f$ and finally obtain $\hat{f}_d$. The final objective of the classification is defined as:

\begin{equation}
    \hat{f}_d = f - DEC_d(ENC_{d}(f) \cdot M_d)
\end{equation}

\begin{equation}
    \label{loss_cls}
    \mathcal{L}_{cls} = -\sum y log(G_{c}(\hat{f}_d)),
\end{equation}

\noindent where $ENC_{d}(\cdot)$ and $DEC_d(\cdot)$ are fully-connected layers for DIM. Because the target domain does not have any annotations, we train $G_c$ with only the source samples. 

\subsection{Semantic Enhancement Module}
\label{subsec:sem}
Although ADG methods are effective in transferring knowledge from the source domain to target domain, their domain discriminators do not focus on foreground features. In particular, these approaches can align the background features that are not important for classification. 

To solve this \textit{background alignment problem}, SEM highlights semantic information for $G_f$ to generate foreground and domain-invariant $f$. Fig. \ref{subfig:sem} shows the structure of SEM. Similar to DIM, we use $M_c \in \mathbb{R}^{N_c \times N_h}$ to effectively align the foreground features of the source and target domains, where $N_c$ is the number of classes. We define class threshold $T_c$, which is the guideline for distinguishing semantic elements within the classifier weight $W_c$. We average $|W_c|$ across the channel dimension and remove unnecessary elements as follows,

\begin{equation}
T_c = \frac{1}{N_h} \sum \sigma(|W_c|)
\end{equation}

\begin{equation}
M_c = I(\sigma|W_c| > T_c).
\end{equation}

Because each row of $M_c$ represents a semantic representation for each class, we cannot directly obtain semantic-enhanced feature $\hat{f}_c$ by multiplying $M_c$ by $f$. We need to change from $M_c$ to batch-based mask $\tilde{M_c}$. To this end, all batch samples require labels. Although source feature $f_s$ has its one-hot vector $y_s$, target feature $f_t$ is fully-unlabeled. Thus, we obtain the target score $p_t$ using $G_c$ ($\sum_{i}^{N_c}p_{t}^{i}=1$). With source labels and target scores, we calculate $\tilde{M_c}$ as,



\begin{equation}
    \tilde{M^{s}_c} = \sum_{i}^{N_c} M^{i}_{c} \cdot y_s^i, \quad \tilde{M^{t}_c} = \sum_{i}^{N_c} M^{i}_{c} \cdot p^{i}_{t} 
\end{equation}

\begin{equation}
    \tilde{M}_c = cat(\tilde{M^{s}_c} | \tilde{M^{t}_c}),
\end{equation}

\noindent where $cat(\cdot)$ is concatenation function. The batch index is omitted. 

To align foreground features in $G_d$, we highlight class-related representations in $f$. Specifically, we obtain $f_c$ by multiplying $\tilde{M_c}$ by $f$ and adding $f_c$ to $f$ as follows:
\begin{equation}
    \hat{f}_c = f + DEC_{c}(ENC_{c}(f) \cdot \tilde{M}_c),
\end{equation}
\noindent where $ENC_{c}(\cdot)$ and $DEC_c(\cdot)$ are fully-connected layers for SEM.

We adapt the adversarial learning framework \cite{ganin2015unsupervised} to our network to enable a two-player minimax game in which $G_d$ tries to distinguish the source and target domains, whereas $G_f$ generates feature vectors that are difficult to differentiate domains. $G_f$ and $G_d$ are connected by the gradient reverse layer that reverses the gradients that flow from $G_d$ to $G_f$. Formally, the objective of adversarial learning $\mathcal{L}_{adv}$ can be written as:
\begin{equation}
 \begin{split}
 \label{loss_adv}
\mathcal{L}_{adv}=\min_{\theta_{G_f}} \max_{\theta_{G_d}, \theta_{\text{C}}} \mathbb{E}_{x_s \sim \mathcal{D}_S} \log G_d(\text{SEM}[G_f(x_s)]) \\
+ \mathbb{E}_{x_t \sim \mathcal{D}_T} \log (1 - G_d(\text{SEM}[G_f(x_t)])),
\end{split}
\end{equation}
where $\theta_{G_{f}}$, $\theta_{G_{d}}$ and $\theta_{\text{c}}$ are the parameters of $G_f$, $G_d$ and SEM respectively.

\subsection{Loss Function}
\label{loss_function}
Our total objective function is composed of two loss terms, one is for the classification loss $\mathcal{L}_{cls}$ (eq. \ref{loss_cls}) and another for the adversarial loss $\mathcal{L}_{adv}$ (eq. \ref{loss_adv}). $\mathcal{L}_{cls}$ removes domain-related features and compares the prediction with the ground-truth for source domain. $\mathcal{L}_{adv}$ highlights class-related features and aligns two foreground domain features. The total loss is defined as:

\begin{equation}
    \mathcal{L} = \mathcal{L}_{cls} + \lambda\cdot\mathcal{L}_{adv},
\end{equation}
where $\lambda$ is trade-off parameter. In this paper, $\lambda=1$.

\begin{table}[]
\centering
\caption{Accuracy (\%) on Office-31 (ResNet-50). MSTN* is a reproduced version by \cite{chang2019domain}. WEMNet and $\text{WEMNet}^ \dagger$ use DANN \cite{ganin2015unsupervised} and MSTN \cite{xie2018learning} as the baselines, respectively.}
\label{tab:exp_office31}
\resizebox{0.5\textwidth}{!}{
\begin{tabular}{ccccccc|c}
\hline
Method    & A$\to$W   & D$\to$W & W$\to$D & A$\to$D & D$\to$A & W$\to$A & Avg  \\ \hhline{========}
ResNet-50 \cite{he2016deep} & 68.4 & 96.7 & 99.3 & 68.9 & 62.5 & 60.7 & 76.1 \\
DANN \cite{ganin2015unsupervised} & 82.0 & 96.9 & 99.1 & 79.7 & 68.2 & 67.4 & 82.2 \\
SimNet & 88.6 & 98.2 & 99.7 & 85.3 & 73.4 & 71.6 & 86.2 \\
MSTN* \cite{xie2018learning} & 91.3 & \textbf{98.9} & \textbf{100.0} & 90.4 & 72.7 & 65.6 & 86.5 \\
CRST \cite{zou2019confidence} & 89.4 & \textbf{98.9} & \textbf{100.0} & 88.7 & 72.6 & 70.9 & 86.8 \\
CDAN+E \cite{long2018conditional} & 94.1 & 98.6 & \textbf{100.0} & \textbf{92.9} & 71.0 & 69.3 & 87.7 \\
TADA \cite{wang2019transferable} & 94.3 & 98.7 & 99.8 & 91.6 & 72.9 & 73.0 & 88.4 \\

\hline
WEMNet    & 93.7 & 98.1 & 99.8 & 86.3 & 75.2 & 69.2 & 87.1   \\
$\text{WEMNet}^ \dagger$ & \textbf{95.6} & 98.1 & \textbf{100.0} & 88.4 & \textbf{75.4} & \textbf{73.9} & \textbf{88.6} \\ 
\hline
\end{tabular}}
\end{table}

\begin{table*}[]
\centering
\caption{Accuracy (\%) on VisDA-2017 (ResNet-101). MSTN* is a reproduced version by \cite{chang2019domain}. WEMNet and $\text{WEMNet}^ \dagger$ use DANN \cite{ganin2015unsupervised} and MSTN \cite{xie2018learning} as the baselines, respectively.}
\label{tab:exp_visda2017}
\resizebox{0.9\linewidth}{!}{%
\begin{tabular}{@{}ccccccccccccc|c@{}}
\toprule
Method & aero & bicycle & bus & car & horse & knife & motor & person & plant & skate & train & truck & Avg \\ \hhline{==============}
ResNet-101 \cite{he2016deep} & 72.3 & 6.1 & 63.4 & \textbf{91.7} & 52.7 & 7.9 & 80.1 & 5.6 & \textbf{90.1} & 18.5 & 78.1 & 25.9 & 49.4 \\
DANN \cite{ganin2015unsupervised} & 81.9 & 77.7 & 82.8 & 44.3 & 81.2 & 29.5 & 65.1 & 28.6 & 51.9 & 54.6 & 82.8 & 7.8 & 57.4 \\
MSTN* \cite{xie2018learning} & 89.3 & 49.5 & 74.3 & 67.6 & 90.1 & 16.6 & \textbf{93.6} & 70.1 & 86.5 & 40.4 & 83.2 & 18.5 & 65.0 \\
MCD \cite{saito2018maximum} & 87.0 & 60.9 & 83.7 & 64.0 & 88.9 & 79.6 & 84.7 & 76.9 & 88.6 & 40.3 & 83.0 & 25.8 & 71.9 \\
ADR \cite{saito2017adversarial} & 87.8 & \textbf{79.5} &  83.7 & 65.3 & 92.3 & 61.8 & 88.9 & 73.2 & 87.8 & 60.0 & 85.5 & 32.3 & 74.8 \\
\hline
WEMNet & \textbf{95.4} & 70.1 & \textbf{84.7} & 51.3 & \textbf{93.2} & \textbf{94.0} & 89.3 & \textbf{77.5} & 88.1 & \textbf{82.8} & 84.1 & \textbf{44.9} & \textbf{75.4} \\
$\text{WEMNet}^ \dagger$ & \textbf{95.4} & 76.1 & 82.0 & 52.3 & 89.8 & \textbf{94.0} & 86.6 & 76.8 & 85.2 & 79.7 & \textbf{86.5} & 41.5  & 74.6 \\ \bottomrule
\end{tabular}%
}
\end{table*}

\section{Experiments}
We conduct extensive evaluations of the two domain adaptation benchmarks, such as Office-31 \cite{saenko2010adapting} and VisDA-2017 \cite{peng2017visda}. We apply the proposed method to popular UDA approaches (DANN \cite{ganin2015unsupervised} and MSTN \cite{xie2018learning}) for comparisons with \textit{state-of-the-art} networks and ablation studies to demonstrate the contributions of our WEMNet.
\subsection{Datasets}
\noindent \textbf{Office-31.} \cite{saenko2010adapting} is the most popular dataset for real-world domain adaptation. Office-31 contains 4,110 images of 31 class in three domains: Amazon (\textbf{A}), Webcam (\textbf{W}) and DSLR (\textbf{D}). We evaluate all methods on six domain adaptation tasks.
\noindent \textbf{VisDA-2017.} \cite{peng2017visda} is a large-scale dataset for syntheticto-real domain adaptation. VisDA-2017 contains 152,397 synthetic images for the source domain and 55,388 real-world images for the target domain.

\subsection{Implementation Details}
In all experiments, we follow the standard protocol for UDA. We use all labeled source data and all unlabeled target data. We use mini-batch stochastic gradient descent (SGD) with a momentum of 0.9, an initial learning rate of 0.01, and a weight decay of 0.001. For Office-31, we use the ResNet-50 \cite{he2016deep} backbone pre-trained on ImageNet \cite{deng2009large} and train the model for 20 epochs. For VisDA-2017, we use ResNet-101 \cite{he2016deep} as the backbone and train for 30 epochs. 

\subsection{Comparison with the state-of-the-art}
\noindent\textbf{Office-31.} In Table \ref{tab:exp_office31}, we compare the proposed methods with recent UDA frameworks on the Office-31 dataset based on ResNet-50. The proposed method demonstrates significant performance improvements. WEMNet, which uses DANN as the baseline, achieves an accuracy of 87.1 \%. Furthermore, $\text{WEMNet}^ \dagger$, which uses MSTN as the baseline, obtains 88.6 \%. Compared to each baseline, WEMNet and $\text{WEMNet}^ \dagger$ increase by 4.9 and 2.1 points, respectively.

\noindent\textbf{VisDA-2017.} To compare with the recent UDA methods on VisDA-2017 dataset based on ResNet-101, we conduct experiments. Table \ref{tab:exp_visda2017} shows that WEMNet achieves remarkable performance compared with other \textit{state-of-the-art} methods. The baseline of WEMNet is DANN, and it achieves an accuracy of 75.4 \%. $\text{WEMNet}^ \dagger$ uses MSTN as the baseline and achieves an accuracy of 74.6 \%. These results show the accuracy improvements of 18.0 and 9.6 points, respectively.  


\begin{table}[]
\centering
\caption{Ablation results of investigating the effects of our components on Office-31.}
\label{tab:exp_components}
\resizebox{0.5\textwidth}{!}{%
\begin{tabular}{c|c|c|c|c|c|c|c|c}
\hline
DIM & SEM & A$\to$W & D$\to$W & W$\to$D & A$\to$D & D$\to$A & W$\to$A & Avg \\ \hline
 &  & 82.0 & 96.9 & 99.1 & 79.7 & 68.2 & 67.4 & 82.2 \\
 \checkmark &  & 91.6 & 96.5 & 98.0 & 84.7 & 74.0 & 67.9 & 85.5 \\
 & \checkmark & 91.7 & 97.0 & 98.0 & 84.9 & 73.6 & 68.6 & 85.6 \\
 \checkmark & \checkmark & \textbf{93.7} & \textbf{98.1} & \textbf{99.8} & \textbf{86.3} & \textbf{75.2} & \textbf{69.2} & \textbf{87.1}  \\ \hline
\end{tabular}%
}
\end{table}

\subsection{Ablation Study}

\begin{table}[]
\centering
\caption{Domain error for source and target features on all pairs of Office-31.}
\label{tab:domain_error}
\resizebox{0.5\textwidth}{!}{%
\begin{tabular}{c|c|c|c|c|c|c|c|c}
\hline
Domain & $f_d$ & A$\to$W  & D$\to$W  & W$\to$D  & A$\to$D  & D$\to$A  & W$\to$A  & Avg  \\ \hline
\multirow{2}{*}{Source} & with & 5.73 & 3.16 & 1.81 & 4.42 & 1.05 & 5.05 & 3.54 \\
                        & without & \textbf{5.19} & \textbf{2.08} & \textbf{1.55} & \textbf{4.07} & \textbf{0.42} & \textbf{3.85} & \textbf{2.86} \\ \hline
\multirow{2}{*}{Target} & with & 0.53 & 7.12 & 2.13 & 4.82 & 7.05 & 5.01 & 4.44 \\
                        & without & \textbf{0.41} & \textbf{6.51} & \textbf{2.00} & \textbf{3.52} & \textbf{6.44} & \textbf{3.93} & \textbf{3.80} \\ \hline
\end{tabular}
}%
\end{table}

Ablation studies are conducted to investigate the effectiveness of the components of the proposed method. Table \ref{tab:exp_components} shows that DIM increases the accuracy by 3.3 points. SEM provides an additional improvement of 3.4 points. These results show that our DIM and SEM are effective in improving classification accuracy.

We also show that DIM can obtain domain-agnostic features in the source and target domains. To this end, we define the domain error $err_{d}$ as follows,

\begin{equation}
    err_{d}(x) = |x-0.5| \times 100.
\end{equation}

To compare $err_{d}$ on Office-31 dataset, we use the trained domain discriminator to obtain domain scores for feature vectors with and without domain-related representation, and measure $err_{d}$ on each score. To minimize the effect of adversarial learning on $err_{d}$, we train the model for 10 epochs, and other hyper-parameters are the same. The results are presented in Table \ref{tab:domain_error}. For all pairs, $err_{d}$ represents a relatively low error for feature vectors without domain-related features.  It represents that DIM can remove domain-related features using domain mask.

\section{Conclusion}
In this paper, we propose a novel and simple WEMNet architecture to improve the generalization performance of the target domain. Based on the concept of Weight-based Mask, we introduce DIM and SEM. To solve the \textit{classification bias problem}, DIM eliminates domain-related features for classification. On the other hand, SEM tackles the \textit{background alignment problem} by enhancing class-related features for domain adaptation. Because we reuse the weights of the domain discriminator and the classifier, we require fewer additional parameters and computations. In addition, the proposed method can easily be applied to any domain discriminator-based UDA approach. We achieve a competitive performance compared with \textit{state-of-the-art} methods on the popular benchmarks. We also conduct ablation studies to demonstrate the effectiveness of the two proposed modules.
\newline\newline
\noindent\textbf{Acknowledgment.} This work was supported by Institute of Information \& communications Technology Planning \& Evaluation(IITP) grant funded by the Korea government(MSIT) (No.2017-0-00897, Development of Object Detection and Recognition for Intelligent Vehicles) and (No.2014-3-00123, Development of High Performance Visual BigData Discovery Platform for Large-Scale Realtime Data Analysis).


\newpage

\bibliographystyle{IEEEbib}
\bibliography{refs}

\begin{thebibliography}{10}

\bibitem{torralba2011unbiased}
Antonio Torralba and Alexei~A Efros,
\newblock ``Unbiased look at dataset bias,''
\newblock in {\em CVPR 2011}. IEEE, 2011, pp. 1521--1528.

\bibitem{ganin2015unsupervised}
Yaroslav Ganin and Victor Lempitsky,
\newblock ``Unsupervised domain adaptation by back propagation,''
\newblock in {\em International conference on machine learning}. PMLR, 2015,
  pp. 1180--1189.

\bibitem{tzeng2017adversarial}
Eric Tzeng, Judy Hoffman, Kate Saenko, and Trevor Darrell,
\newblock ``Adversarial discriminative domain adaptation,''
\newblock in {\em Proceedings of the IEEE conference on computer vision and
  pattern recognition}, 2017, pp. 7167--7176.

\bibitem{zhu2022localized}
Wei Zhu, Le~Lu, Jing Xiao, Mei Han, Jiebo Luo, and Adam~P Harrison,
\newblock ``Localized adversarial domain generalization,''
\newblock in {\em Proceedings of the IEEE/CVF Conference on Computer Vision and
  Pattern Recognition}, 2022, pp. 7108--7118.

\bibitem{quinonero2008covariate}
Joaquin Qui{\~n}onero-Candela, Masashi Sugiyama, Anton Schwaighofer, and
  N~Lawrence,
\newblock ``Covariate shift and local learning by distribution matching,''
  2008.

\bibitem{long2017deep}
Mingsheng Long, Han Zhu, Jianmin Wang, and Michael~I Jordan,
\newblock ``Deep transfer learning with joint adaptation networks,''
\newblock in {\em International conference on machine learning}. PMLR, 2017,
  pp. 2208--2217.

\bibitem{xie2018learning}
Shaoan Xie, Zibin Zheng, Liang Chen, and Chuan Chen,
\newblock ``Learning semantic representations for unsupervised domain
  adaptation,''
\newblock in {\em International conference on machine learning}. PMLR, 2018,
  pp. 5423--5432.

\bibitem{saenko2010adapting}
Kate Saenko, Brian Kulis, Mario Fritz, and Trevor Darrell,
\newblock ``Adapting visual category models to new domains,''
\newblock in {\em European conference on computer vision}. Springer, 2010, pp.
  213--226.

\bibitem{peng2017visda}
Xingchao Peng, Ben Usman, Neela Kaushik, Judy Hoffman, Dequan Wang, and Kate
  Saenko,
\newblock ``Visda: The visual domain adaptation challenge,''
\newblock {\em arXiv preprint arXiv:1710.06924}, 2017.

\bibitem{chang2019domain}
Woong-Gi Chang, Tackgeun You, Seonguk Seo, Suha Kwak, and Bohyung Han,
\newblock ``Domain-specific batch normalization for unsupervised domain
  adaptation,''
\newblock in {\em Proceedings of the IEEE/CVF conference on Computer Vision and
  Pattern Recognition}, 2019, pp. 7354--7362.

\bibitem{he2016deep}
Kaiming He, Xiangyu Zhang, Shaoqing Ren, and Jian Sun,
\newblock ``Deep residual learning for image recognition,''
\newblock in {\em Proceedings of the IEEE conference on computer vision and
  pattern recognition}, 2016, pp. 770--778.

\bibitem{zou2019confidence}
Yang Zou, Zhiding Yu, Xiaofeng Liu, BVK Kumar, and Jinsong Wang,
\newblock ``Confidence regularized self-training,''
\newblock in {\em Proceedings of the IEEE/CVF International Conference on
  Computer Vision}, 2019, pp. 5982--5991.

\bibitem{long2018conditional}
Mingsheng Long, Zhangjie Cao, Jianmin Wang, and Michael~I Jordan,
\newblock ``Conditional adversarial domain adaptation,''
\newblock {\em Advances in neural information processing systems}, vol. 31,
  2018.

\bibitem{wang2019transferable}
Ximei Wang, Liang Li, Weirui Ye, Mingsheng Long, and Jianmin Wang,
\newblock ``Transferable attention for domain adaptation,''
\newblock in {\em Proceedings of the AAAI Conference on Artificial
  Intelligence}, 2019, vol.~33, pp. 5345--5352.

\bibitem{saito2018maximum}
Kuniaki Saito, Kohei Watanabe, Yoshitaka Ushiku, and Tatsuya Harada,
\newblock ``Maximum classifier discrepancy for unsupervised domain
  adaptation,''
\newblock in {\em Proceedings of the IEEE conference on computer vision and
  pattern recognition}, 2018, pp. 3723--3732.

\bibitem{saito2017adversarial}
Kuniaki Saito, Yoshitaka Ushiku, Tatsuya Harada, and Kate Saenko,
\newblock ``Adversarial dropout regularization,''
\newblock {\em arXiv preprint arXiv:1711.01575}, 2017.

\bibitem{deng2009large}
Jia Deng,
\newblock ``A large-scale hierarchical image database,''
\newblock {\em Proc. of IEEE Computer Vision and Pattern Recognition, 2009},
  2009.

\end{thebibliography}

\end{document}